\documentclass{llncs}
\usepackage{llncsdoc}
 \usepackage{epsfig}
 \usepackage{graphicx}
 \usepackage{verbatim}
 \usepackage{color}

\begin{document}
\newcommand{\matrixx}{\cal{X}}
\newcommand{\vecx}{\vec{x}}
\newcommand{\vecX}{\vec{X}}
\newcommand{\mydefine}{\stackrel{def}{=}}
\newcommand{\mycolor}{} 
\newcommand{\melf}{\phi_{f}}
\newcommand{\linearf}{l_{f}}
\newcommand{\nspeech}{y}
\newcommand{\N}{{\cal N}}
\newcommand{\myalpha}{\xi}
\newcommand{\mymark}{}
\newcommand{\myend}{$$}
\newcommand{\speak}[1]{{/{\em #1}/}}
\newcommand{\word}[1]{{{\em #1}}}
\newcommand{\n}{{\cal N}}

\def\x{{\mathbf x}}
\def\L{{\cal L}}
\let\remark=\comment
\let\endremark=\endcomment

\title{Evaluating the Performance of a Speech Recognition based System}
%
\author{Vinod Kumar Pandey,  Sunil Kumar Kopparapu}
%
\institute{TCS Innovation Labs - Mumbai, \\ Tata Consultancy Services,
Pokharan Road 2 \\ Thane 400 601, Maharastra, India. \\ \{vinod.pande, sunilkumar.kopparapu\}@tcs.com}
%
%
%
\maketitle

\begin{abstract}

Speech based solutions have taken center stage with growth in the services
industry where there is a need to cater to a very large number of people from all 
strata of the society. While natural language speech interfaces are the talk in 
the research community, yet in practice, menu based speech solutions thrive. 
Typically in a menu based speech solution the user is required to respond by 
speaking from a closed set of words when prompted by the system.
A sequence of 
human speech response to the IVR prompts results in the completion of a 
transaction. A transaction is deemed successful if the speech solution can 
correctly recognize all the spoken utterances of the user whenever prompted
by 
the system. The usual mechanism to evaluate the performance of a speech 
solution is to do an extensive test of the system by putting it to actual {\em 
people use} and then evaluating the performance by analyzing the logs 
for successful transactions. 
This 
kind of evaluation could lead to 
dissatisfied test users especially if the performance of the 
system were to result in 
a poor transaction completion rate. To negate this 
the Wizard of Oz approach is adopted during evaluation of a speech system. 
Overall this kind of evaluations is an expensive 
proposition both in terms of time and cost.
In this paper, we propose a method to evaluate the 
performance of a speech solution without actually putting it to {\em people use}. 
We first describe the methodology and then show experimentally that this can be 
used to identify the performance bottlenecks of the speech solution even before 
the system is actually used thus saving evaluation time and expenses.

\end{abstract}
\begin{keywords}
Speech solution evaluation, Speech recognition,  Pre-launch recognition performance measure.
\end{keywords}
\section{Introduction}
\label{sec:intro}
\begin{remark}
In India, (a) a large percentage of the population is not English literate; (b) 
they can speak their language but might not be able to write or read the same 
language; (c) people have access to mobile phone thanks to the proliferation of 
mobile phones in the last couple of years; (d) no major access to broadband or 
Internet or computers; (e) sending SMS in local Indian language is still 
cumbersome even to the people who can write in their language; and (f) majority 
of this population are in need of some transactional information (travel inquiry, 
news, stock quotes, yellow pages etc) or the other very often. In total sum, in 
the Indian context, there exists a huge opportunity for speech based solutions 
especially if they can address usage by masses. Speech based solutions require no 
marketing and will catch like wild fire if the transaction completion rate is 
high. The success of a speech solution lies in being able to build an automatic 
speech recognition (ASR) based system which is robust, works for large 
populations \cite{sun10}, \cite{lua10}, \cite{zhao10}, \cite{kim10} and is able 
to service the masses in Indian languages. The challenge however lies in the fact 
that there are too many languages, even more dialects that makes even a menu 
based speech solution untamed in the Indian scenario.
\end{remark}
There are several commercial menu based ASR systems available around the world for 
a significant number of languages and interestingly speech solution based on 
these ASR are being used with good success in the Western part of the globe 
\cite{sun10}, \cite{lua10}, \cite{zhao10}, \cite{kim10}. 
Typically, a menu based ASR system restricts user to speak from a pre-defined 
closed set of words for enabling a transaction. 
\begin{remark} However, these commercially 
viable menu based ASR systems can not be directly plugged into the Indian 
scenarios and for Indian languages for mass use because of several idiosyncrasies 
of the Indian conditions. The case in point is the observation that though speech 
solution is an apt solution in the Indian context, there is no speech solution of 
any reasonable repute in deployment for Indian languages. Some of the reasons for 
this are (a) there are officially 22 different languages in India and a larger 
variety of dialect variants; (b) Indian languages have not been as richly studied 
compared to the western languages especially English; (c) there is a large 
variation in accents of the speakers; (d) it is very difficult to fine tune and 
deploy ASR system developed primarily for western languages in India; (e) Indian 
telephone lines are far more noisy compared to the Western counterparts. This 
leads to the situation where using the ASR system, developed for Western 
languages and conditions, for Indian conditions and languages needs significant 
innovation and changes to make it suitable for use in India. Variations in the 
dialects and accents, high cost associated with creating the speech corpora etc 
pose practical difficulties in developing an speech recognition system for use in 
India. This large diversity in languages and the large number of speech solution 
application domains (travel inquiry, news, stock quotes, yellow pages etc) calls 
for a strategy which can provide a workable system.
 There has been continuous effort in developing speech recognition for  different Indian languages to address different domains \cite{kumar04},
\cite{banerjee08}, \cite{thangarajan08}, \cite{patel09}, \cite{sharma08}, \cite{agarwal10}. 
\end{remark}
Before commercial deployment of a speech solution it is imperative to have a quantitative measure of the performance of the speech 
 solution which is primarily based on the speech recognition accuracy of the 
speech engine used. Generally, the recognition performance of any speech recognition 
based solution is quantitatively evaluated by putting it to actual use by the 
people who are the intended users and then analyzing the logs to identify 
successful and unsuccessful transactions. This evaluation is then used to 
identifying any further improvement in the speech recognition based 
solution to better the overall 
transaction completion rates. This process of evaluation is both time consuming 
and expensive. For evaluation one needs to identify a set of users and also 
identify the set of actual usage situations and perform the test. It is also 
important that the set of users are able to use the system with ease meaning that 
even in the test conditions the performance of the system, should be good, while 
this can not usually be guaranteed this aspect of keeping the user experience 
good makes it necessary to employ a wizard of Oz (WoZ) approach. Typically this 
requires a human agent in the loop during actual speech transaction where the 
human agent corrects any mis-recognition by actually listening to the 
conversation between the human user and the machine without the user knowing that 
there is a human agent in the loop. The use of WoZ is another expense in the 
testing a speech solution. All this makes testing a speech solution an expensive 
and time consuming procedure.

 In this paper, we describe a method to evaluate the performance of a speech 
solution without actual people using the system as is usually done. We then show 
how this method was adopted to evaluate a speech recognition based 
solution as a case study. This 
is the main contribution of the paper. 
The rest of the paper is organized as follows. The method for evaluation 
without testing is described in Section \ref{sec:two}. In Section 
\ref{sec:three} we present a case study and conclude
 in Section \ref{sec:four}.

\section{Evaluation without Testing}%
\label{sec:two}

\begin{figure}
\centerline{
\includegraphics[width=0.95\textwidth]{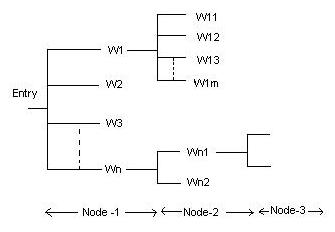}
}
\caption{Schematic of a typical menu based ASR system ($W_n$ is spoken word).
}
 \label{fig:nodes}
\end{figure}
Fig. \ref{fig:nodes} shows the schematic of a typical menu based speech solution 
having $3$ nodes. At each node there are a set of words that the user is expected 
to speak and the system is supposed to recognize. In this particular schematic, 
at the entry node the user can speak any of the { $n$} words, namely $W_1$ or 
$W_2$ or $\cdots$ or $W_n$; $n$ is usually called the perplexity of the node 
in the speech 
literature. The larger the $n$ the more the perplexity and higher the confusion 
and hence lower the recognition accuracies. In most commercial 
speech solutions the 
perplexity is kept very low, typically a couple of words. Once the word at the 
entry node has been recognized (say word $W_k$ has been recognized), the system 
moves on to the second node where the active list of words to be recognized could 
be one of $W_{k1}$, $W_{k2}$, $W_{k3}$, ... $W_{kp}$ if the perplexity at the 
$k^{th}$ node is {$p$}. This is carried on to the third node. A transaction is 
termed successful {\em if and only if} the recognition at each of the three nodes is 
correct.  For example, typically in a banking speech solution the entry node 
could expect someone to speak among \speak{credit card}\footnote{We will use 
/ / to indicate the spoken word. For example /$W_1$/ represents the spoken 
equivalent of the written word $W_1$.}, \speak{savings account}, 
\speak{current account}, \speak{loan product}, \speak{demat}, and \speak{mutual 
fund transfer} which has a perplexity of $6$. Once a person speaks, say,
\speak{savings 
account} and is recognized correctly by the system, 
at the second node it could be \speak{account 
balance} or \speak{cheque} or \speak{last $5$ transactions} 
(perplexity $3$) and at the 
third node (say, on recognition of \speak{cheque}) it could be \speak{new cheque book 
request}, \speak{cheque status}, and \speak{stop cheque request} (perplexity 
$3$).  
\begin{note}
Though we will not dwell on this, it is important to note that an error in 
recognition at the entry node is more expensive than a recognition error at a 
lower node. 
\end{note}
\begin{figure}
\begin{center}
\setlength{\unitlength}{3947sp}%
\begingroup\makeatletter\ifx\SetFigFont\undefined%
\gdef\SetFigFont#1#2#3#4#5{%
  \reset@font\fontsize{#1}{#2pt}%
  \fontfamily{#3}\fontseries{#4}\fontshape{#5}%
  \selectfont}%
\fi\endgroup%
\begin{picture}(5744,3292)(3579,-3931)
\thicklines
{\color[rgb]{0,0,0}\put(7501,-1411){\framebox(1500,750){}}
}%
{\color[rgb]{0,0,0}\put(5701,-3436){\framebox(1200,675){}}
}%
\thinlines
{\color[rgb]{0,0,0}\put(5101,-3061){\vector( 1, 0){600}}
}%
{\color[rgb]{0,0,0}\put(8101,-1411){\line( 0,-1){825}}
\put(8101,-2236){\line(-1, 0){1800}}
\put(6301,-2236){\vector( 0,-1){525}}
}%
{\color[rgb]{0,0,0}\put(6301,-3436){\vector( 0,-1){375}}
}%
\thicklines
{\color[rgb]{0,0,0}\put(7801,-2761){\framebox(1200,300){}}
}%
{\color[rgb]{0,0,0}\put(7801,-3436){\framebox(1500,375){}}
}%
\thinlines
{\color[rgb]{0,0,0}\put(7801,-2536){\line(-1, 0){300}}
\put(7501,-2536){\line( 0,-1){750}}
\put(7501,-3286){\line( 1, 0){300}}
}%
{\color[rgb]{0,0,0}\put(7501,-2536){\line(-1, 0){900}}
\put(6601,-2536){\vector( 0,-1){225}}
}%
\thicklines
{\color[rgb]{0,0,0}\put(3601,-3511){\framebox(1500,675){}}
}%
\put(7801,-1111){\makebox(0,0)[lb]{\smash{{\SetFigFont{10}{12.0}{\rmdefault}{\mddefault}{\updefault}{\color[rgb]{0,0,0}Acoustic Models}%
}}}}
\put(6451,-3886){\makebox(0,0)[lb]{\smash{{\SetFigFont{10}{12.0}{\rmdefault}{\mddefault}{\updefault}{\color[rgb]{0,0,0}Text Output}%
}}}}
\put(7951,-2686){\makebox(0,0)[lb]{\smash{{\SetFigFont{10}{12.0}{\rmdefault}{\mddefault}{\updefault}{\color[rgb]{0,0,0}Lexicon}%
}}}}
\put(7876,-3211){\makebox(0,0)[lb]{\smash{{\SetFigFont{10}{12.0}{\rmdefault}{\mddefault}{\updefault}{\color[rgb]{0,0,0}Language Model}%
}}}}
\put(3826,-3211){\makebox(0,0)[lb]{\smash{{\SetFigFont{10}{12.0}{\rmdefault}{\mddefault}{\updefault}{\color[rgb]{0,0,0}Speech}%
}}}}
\put(6001,-2986){\makebox(0,0)[lb]{\smash{{\SetFigFont{10}{12.0}{\rmdefault}{\mddefault}{\updefault}{\color[rgb]{0,0,0}Speech}%
}}}}
\put(5776,-3211){\makebox(0,0)[lb]{\smash{{\SetFigFont{10}{12.0}{\rmdefault}{\mddefault}{\updefault}{\color[rgb]{0,0,0}Recognition}%
}}}}
\end{picture}%
\end{center}
\caption{A typical speech recognition system. In a menu based system the
language model is typically the set of words that need to be recognized at a
given node.}
\label{fig:sr}
\end{figure}
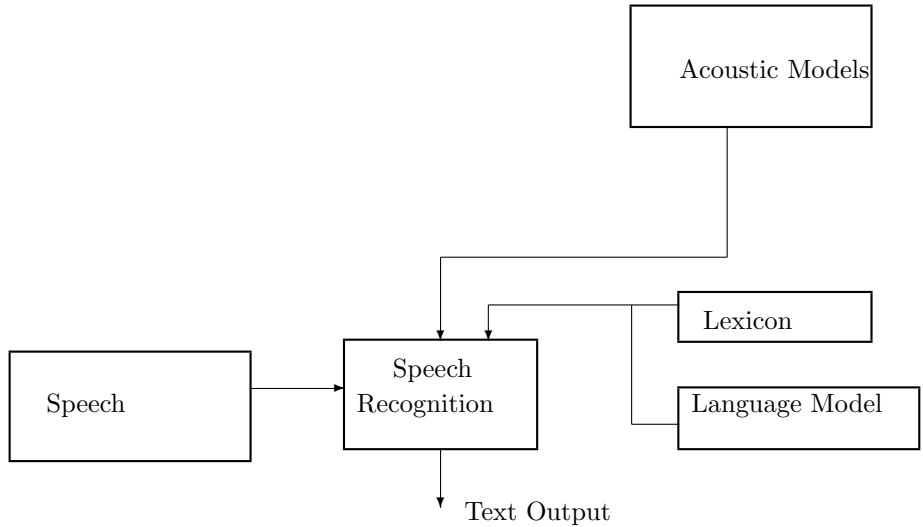
Based on the call flow, and the domain the system can have several 
nodes for completion of a transaction. Typical menu based speech solutions 
strive for a $3$ - $5$ level nodes to make it usable.  
In any speech based solution (see Fig. \ref{fig:sr})
first the spoken utterance is hypothesized into a sequence of phonemes using
the acoustic models. Since the phoneme recognition accuracy is low, 
instead of choosing one phoneme
it identifies l-best (typically $l = 3$) matching phonemes.
This phone lattice is then matched with all the expected 
words (language model) at that node to find the best match. 
For a node with perplexity $n$ the 
constructed phoneme lattice of the spoken utterance is compared with the phoneme 
sequence representation of  all the {$n$} words (through the lexicon
which is one of he key components of a speech recognition system). 
The hypothesized phone lattice is 
declared one of the {$n$} words depending on the closeness of the phoneme 
lattice to the phoneme representation
of the {$n$} words.

We hypothesize that we can identify the performance of a menu based speech 
system by identifying the possible confusion among all the words that are 
active at a given node. \begin{note}If active words at a given node are phonetically similar 
it becomes difficult for the speech recognition system to distinguish them which 
in turn leads to recognition errors.\end{note}
 We used Levenshtein distance 
\cite{gusfield01}, \cite{navarro01} a well known measure to analyze and identify 
the confusion among the active words at a given node.  This analysis gives a list 
of all set of words that have a high degree of confusability among them; 
this understanding can be then 
used to (a) restructure the set of active words at that node and/or (b) train the 
words that can be confused by using a larger corpus of speech data. This allows 
the speech recognition engine to be equipped to be able to distinguish the 
confusing words better. 
Actual use of this analysis was carried out for a 
speech solution developed for Indian Railway Inquiry System 
to identify bottlenecks in the 
system before its actual launch. 

\section{Case Study}
 \label{sec:three}

 \begin{figure}
\centerline{ 
\includegraphics[width=1.0\textwidth]{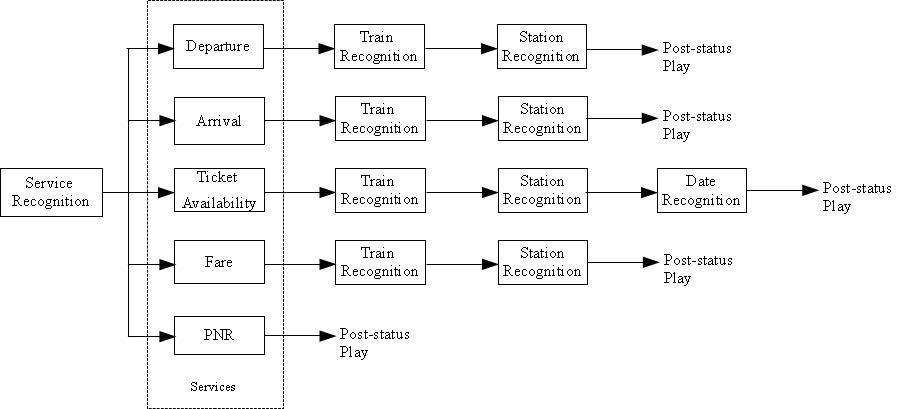}
}
\caption{Call flow of Indian Railway Inquiry System ($W_n$ is spoken word)
}
\label{fig:nodes1}
\end{figure}

A schematic of a speech based Railway Information system, developed for Hindi 
language is shown in Fig. \ref{fig:nodes1}. The system enables user to get 
information on five different services, namely, (a) {\sc Arrival} of a given 
train at a given station, (b) {\sc Departure} of a given train at a given 
station, (c) {\sc Ticket availability} on a given date in a given train between 
two stations, and class, (d) {\sc Fare} in a given class in a given train between 
two stations, and (e) {\sc PNR status}. At the first recognition node (node-1), 
there are one or more active words corresponding to each of these services. For 
example, for selecting the service {\sc Fare}, the user can speak among 
\speak{kiraya jankari}, \speak{kiraya}, \speak{fare}. 
Similarly, for selecting service {\sc 
Ticket availability}, user can speak \speak{upalabdhata jankari} or \speak{ticket 
availability} or \speak{upalabdhata}. 
\begin{note}
Generally the perplexity at a node is greater than on equal to the number of
words that need to be recognized at that node.
\end{note}
In this manner each of the services could 
have multiple words or phrases that can mean the same thing and the speaker could 
utter any of these words to refer to that service. The sum of all the possible 
different ways in which a service can be called ($d_{i}$) summed over all the
$5$
services gives the perplexity ({$\n$}) at that 
node, namely, 
 \begin{equation}
 \n=\sum_{i=0}^{5}d_{i}
 \label{eq:per}
 \end{equation}
The speech recognition engine matches the phoneme lattice of the spoken
utterance with all 
the {$\n$} words which are active. 
The active word (one among the {
$\n$} words) with highest likelihood score is the recognized word. In order to avoid 
low likelihood recognitions a threshold is set so that even the best likelihood
word
is returned only if the likelihood score is greater than the predefined 
threshold.  Completion of a service requires recognitions at several nodes with 
different perplexity at each node.  Clearly depending on the type of service that 
the user is wanting to use; the user has to go through different number of 
recognition 
nodes. For example, to complete the {\sc Arrival} service it is required to pass 
through $3$ recognition nodes namely (a) selection of a service, (b) selection of a 
train name and (c) selection of the railway station. While the perplexity (the 
words that are active) at the service selection node is fixed the perplexity at 
the station selection node could depend on the selection of the train name at 
an earlier node.  For example, if the selected train stops at $23$ stations, then 
the perplexity at the station selection node will be 
$\ge 23$. 
\begin{remark}
However, 
if train stops at a fewer number of stations, then the perplexity at the station 
selection node is less than $23$\footnote{Note that a station name can be spoken 
in several ways. For example, '{\em Mumbai Chhatrapati Shivaji Terminus}' can be 
spoken as '{\em CST}' or '{\em Mumbai}' or '{\em Bombay}' or '{\em VT}' or '{\em 
Chhatrapati Shivaji Terminus}' or '{\em Mumbai Chhatrapati Shivaji Terminus}' or 
'{\em Victoria Terminus}' or '{\em Bombay VT}'.} .
\end{remark}

 For confusability analysis at each of the node, we have used the Levenshtein 
distance \cite{navarro01} or the edit distance as is well known in computer 
science literature. We found that the utterances \speak{Sahi}
and \speak{Galat} have $100$\% recognition. 
These words {Sahi} is represented by the string of phonemes in the lexicon as 
{\sc S AA HH I} and the word {Galat} is represented as the phoneme sequence
{\sc G L AX tT} in the lexicon.  
We identified the edit distance between these two words Sahi 
and Galat and used that distance measure as the 
threshold that is able to differentiate any two words (say 
$T$). So if the distance between any two active words at a given recognition node 
is lower than the threshold $T$, then there is a greater chance that those two 
active words could get confused (one word could be recognized as the other which 
is within a distance of $T$). There are ways in which this possible 
misrecognition words could be avoided.  The easiest way is to make sure that 
these two words together are not active at a given recognition node.

\begin{table}
\caption{List of Active Words at node 1}
\label{tab:active_words} 
\begin{center}
    \begin{tabular}{ | l || l | l |} \hline
    W. No. & Active Word & Phonetic \\ \hline \hline
	${W}_1$ & kiraya\_jankari & K I R AA Y AA J AA tN K AA R I \\ \hline
	${W}_2$ & kiraya & K I R AA Y AA \\ \hline
	${W}_3$ & fare & F AY R \\ \hline
	${W}_4$ & aagaman\_jankari & AA G AX M AX tN J AA tN K AA R I \\ \hline
	${W}_5$ & aagaman & AA G AX M AX tN \\ \hline
	${W}_6$ & arrival\_departure & AX R AA I V AX L dD I P AA R CH AX R \\ \hline
	${W}_7$ & upalabdhata\_jankari & U P AX L AX B tDH AX tT AA J AA tN K AA R I \\ \hline
	${W}_8$ & ticket\_availability & tT I K EY tT AX V AY L AX B I L I tT Y \\ \hline
	${W}_9$ & upalabdhata & U P AX L AX B tD AX tTH AA \\ \hline
	${W}_{10}$ & arrival & AX R AA I V AX L \\ \hline
	${W}_{11}$ & prasthan & P R AX S tTH AA tN \\ \hline
	${W}_{12}$ & departure & dD I P AA R CH AX R \\ \hline
	${W}_{13}$ & p\_n\_r\_jankari & P I EY tN AA R J AA tN K AA R I \\ \hline
	${W}_{14}$ & p\_n\_r & P I AX tN AA R \\ \hline \hline
    \end{tabular}
\end{center}
\end{table}

Table \ref{tab:active_words} shows the list of active word at the node 1 
when the speech application was initially designed and Table 
\ref{tab:matrix} shows the edit distance between all the active words at the node 
service given in Fig. \ref{fig:nodes1}. 
The distance between words \word{Sahi} and \word{ 
Galat} was found to be $5.7$ which was set at the threshold, namely $T = 5.7$. 
This threshold 
value was used to identify confusing active words. Clearly, as seen in the Table 
\label{tab:two} the distance between word pairs \word{fare}, \word{pnr} and
\word{ 
pnr}, \word{prasthan} is $5.2$ and $5.8$ respectively, which is very close to 
the threshold value of $5.7$. This can cause a high possibility that \speak{fare} 
may get recognized as \word{pnr} and vice-versa. 
 \begin{table}
\caption{Distance Measurement for Active Words at  Node 1 of the Railway
Inquiry System }
\label{tab:matrix} 
\begin{center}
    \begin{tabular}{ | l || l | l | l | l | l | l | l | l | l | l | l | l | l | l |} \hline
     & $W_1$ & $W_2$ & $W_3$ & $W_4$ & $W_5$ & $W_6$ & $W_7$ & $W_8$ & $W_9$ &
$W_{10}$ & $W_{11}$ & $W_{12}$ & $W_{13}$ & $W_{14}$ \\ \hline \hline
	$W_1$ & 0 & 8.4 & 14.2 & 8.5 & 14.1 & 15.3 & 11 & 20 & 17.1 & 13 & 13 & 13.2 & 6.2 & 11.2 \\ \hline
	$W_2$ & 8.4 & 0 & 7.2 & 13.8 & 8.5 & 14.7 & 17.8 & 15.7 & 11 & 7.2 & 7.8 & 7.7 & 11.2 & 6.2 \\ \hline
	$W_3$ & 14.2 & 7.2 & 0 & 14.2 & 7.2 & 14.8 & 18.2 & 15.8 & 11.2 & 8.2 & 8.2 & 7.8 & 14.2 & \textbf{5.8} \\ \hline
	$W_4$ & 8.5 & 13.8 & 14.2 & 0 & 8.4 & 15.9 & 9.6 & 19.7 & 14.3 & 13 & 13 & 14.7 & 8.2 & 11.2 \\ \hline
	$W_5$ & 14.1 & 8.5 & 7.2 & 8.4 & 0 & 13.2 & 16.7 & 15.7 & 9.7 & 7.2 & 6.7 & 8.2 & 14.1 & 7.1 \\ \hline
	$W_6$ & 15.3 & 14.7 & 14.8 & 15.9 & 13.2 & 0 & 18.7 & 18.9 & 15.5 & 9.4 & 13.7 & 7 & 17.3 & 11.8 \\ \hline
	$W_7$ & 11 & 17.8 & 18.2 & 9.7 & 16.7 & 18.7 & 0 & 20 & 11.2 & 17.1 & 15.6 & 17.8 & 10.2 & 13.8 \\ \hline
	$W_8$ & 20 & 15.7 & 15.8 & 19.7 & 15.7 & 18.9 & 20 & 0 & 14.5 & 15.8 & 17.5 & 16.5 & 18.6 & 15.7 \\ \hline
	$W_9$ & 17.1 & 11 & 11.2 & 14.3 & 9.7 & 15.5 & 11.2 & 14.5 & 0 & 10 & 9.2 & 11.1 & 16.3 & 9.7 \\ \hline
	$W_{10}$ & 13 & 7.2 & 8.2 & 13.1 & 7.2 & 9.4 & 17.1 & 15.8 & 10 & 0 & 8.5 & 8 & 13 & 7.8 \\ \hline
	$W_{11}$ & 13.1 & 7.8 & 8.2 & 13.1 & 6.7 & 13.7 & 15.7 & 17.5 & 9.2 & 8.5 & 0 & 8.4 & 11.7 & \textbf{5.2} \\ \hline
	$W_{12}$ & 13.2 & 7.7 & 7.8 & 14.7 & 8.2 & 7 & 17.8 & 16.5 & 11.1 & 8.1 & 8.4 & 0 & 12.1 & 6.8 \\ \hline
	$W_{13}$ & 6.2 & 11.2 & 14.2 & 8.2 & 14.1 & 17.3 & 10.2 & 18.6 & 16.3 & 13.1 & 11.7 & 12.1 & 0 & 9.8 \\ \hline
	$W_{14}$ & 11.2 & 6.2 & \textbf{5.8} & 11.2 & 7.1 & 11.8 & 13.8 & 15.7 & 9.6
& 7.8 & \textbf{5.2} & 6.8 & 9.8 & 0  \\ \hline \hline
    \end{tabular}
\end{center}
\end{table}
One can derive from the analysis of the active words that \word{ 
fare} and \word{pnr} can not coexist as active words at the same node. 
The 
result of the analysis was to remove the active words \word{fare} and
\word{pnr} at that node. 

When the speech system was 
actually tested by giving speech samples, $17$ out of $20$ instances of
 \speak{pnr} was 
was recognized as \word{fare} and vice-versa. Similarly $19$ out of 
$20$ instances \speak{pnr} was misrecognized as \word{prasthan} and vice versa 
This confusion is expected as can be seen from the edit distance analysis of the 
active words in the Table \ref{tab:matrix}.
This modified active word list (removal of \word{fare} and 
\word{pnr}) increased the recognition accuracy at the service 
node (Fig. \ref{fig:nodes1}) by as much as $90$\%. 

A similar analysis was carried 
out at other recognition nodes and the active word list was suitably 
modified to avoid possible confusion between active word pair. This analysis 
and modification of the list of active words at a node resulted 
in a significant improvement in the transaction completion rate.
We will present more experimental results in the final paper.

 \section{Conclusion}
 \label{sec:four}

In this paper we proposed a methodology to identify words that could lead to
confusion at any given node of a speech recognition based system. We 
used edit distance as the metric to identifying the 
possible confusion between the active words.
We 
showed that this metric can be used effectively to enhance the performance of a 
speech solution without actually putting it to people test. There is a 
significant saving in terms of being able to identify recognition bottlenecks in 
a menu based speech solution through this analysis because it does not require 
actual people testing the system. This methodology was adopted to restructuring 
the set of active words at each node for better speech recognition in an actual 
menu based speech recognition system that caters to masses.


\end{document}